\documentclass[10pt,twocolumn,letterpaper]{article}
\usepackage{cvpr}              
\usepackage{multirow}
\usepackage{float}
\usepackage{nameref}
\usepackage{bbding}
\usepackage{wasysym}
\usepackage{amssymb}

\usepackage{xcolor}
\usepackage{fontawesome}
\definecolor{lightcoral}{RGB}{240,128,128}
\definecolor{skyblue}{RGB}{135,206,235}
\definecolor{orchid}{RGB}{218,112,214}

\definecolor{cvprblue}{rgb}{0.21,0.49,0.74}

\usepackage[pagebackref, colorlinks, linkcolor=red, citecolor=green , filecolor=magenta, urlcolor=cvprblue]{hyperref}

\def\paperID{4799} 
\def\confName{CVPR}
\def\confYear{2025}

\title{SocialMOIF: Multi-Order Intention Fusion for Pedestrian Trajectory Prediction}

\author{
	Kai Chen$^{1*}$, 
	Xiaodong Zhao\text{(\faEnvelopeO)}$^{1*}$, 
	Yujie Huang$^{1}$, 
	Guoyu Fang$^{1}$, \\
	Xiao Song$^{2}$, 
	Ruiping Wang$^{3}$, 
	Ziyuan Wang$^{1}$\\
	$^{1}$ Nanjing University of Aeronautics and Astronautics \\
	$^{2}$ Beihang University 
	$^{3}$ Nanyang Technological University\\
    {\tt\small \{chen\_kai, xdzhao, yj\_huang, fanggy, zy\_wang\}@nuaa.edu.cn}\\
    {\tt\small songxiao@buaa.edu.cn, ruiping.wang@ntu.edu.sg}
}

\begin{document}
\maketitle
\renewcommand{\thefootnote}{} 
\footnotetext{* Equal contribution.}
\footnote{Code is available at \href{https://github.com/XiaodZhao/SocialMOIF}{\textbf{https://github.com/XiaodZhao/SocialMOIF}}.}

\begin{abstract}
    The analysis and prediction of agent trajectories are crucial for decision-making processes in intelligent systems, with precise short-term trajectory forecasting being highly significant across a range of applications. Agents and their social interactions have been quantified and modeled by researchers from various perspectives; however, substantial limitations exist in the current work due to the inherent high uncertainty of agent intentions and the complex higher-order influences among neighboring groups. SocialMOIF is proposed to tackle these challenges, concentrating on the higher-order intention interactions among neighboring groups while reinforcing the primary role of first-order intention interactions between neighbors and the target agent. This method develops a multi-order intention fusion model to achieve a more comprehensive understanding of both direct and indirect intention information. Within SocialMOIF, a trajectory distribution approximator is designed to guide the trajectories toward values that align more closely with the actual data, thereby enhancing model interpretability. Furthermore, a global trajectory optimizer is introduced to enable more accurate and efficient parallel predictions. By incorporating a novel loss function that accounts for distance and direction during training, experimental results demonstrate that the model outperforms previous state-of-the-art baselines across multiple metrics in both dynamic and static datasets.
\end{abstract}    

\section{Introduction}
\begin{figure}[ht!]
  \centering 
  \includegraphics[scale=0.37]{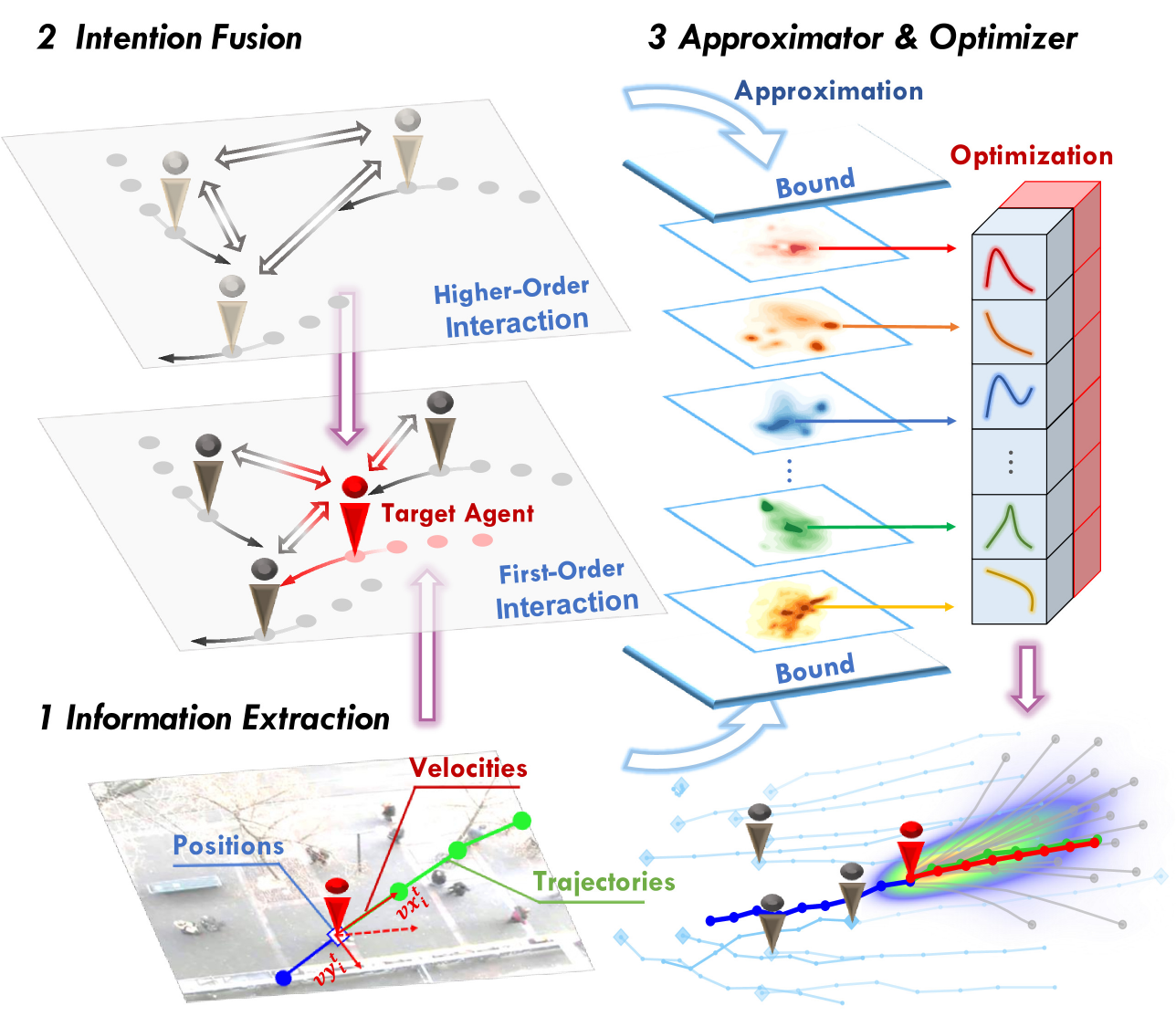}
  \caption{The process of SocialMOIF includes: (1) extracting the position, velocity and trajectory information of agents, (2) achieving multi-order fused intention modeling, (3) guiding the distribution of trajectories and performing global optimization. Finally we obtain accurate parallel predictions.}
  \label{fig1}
\end{figure}
Trajectory prediction is of considerable importance for intelligent transportation and surveillance systems. Studies indicate that trajectories derived from the physical world encompass rich spatiotemporal intention information \cite{1,4,11,16,33}. Therefore, a significant challenge arises in effectively capturing (data modeling) and analyzing (encoding-decoding) processes related to the complex and diverse historical social intentions of agents, while linking this information to future trajectory predictions for enhanced efficiency.

\textbf{History social intentions.} Existing research can be broadly classified into two main categories: \textbf{knowledge-driven methods} primarily depend on expert knowledge. For instance, biomimetic method \cite{3} achieved agent localization and intention interaction modeling by simulating the positioning of marine animals. In contrast, certain studies utilized game-theoretic approaches to analyze social intentions through the simulation of agent flows \cite{52,55} and evacuation processes \cite{49,51}, relying on explicit rules or formulas. However, the behavioral patterns and interaction rules among agents in social scenarios displayed high nonlinearity and uncertainty \cite{7,18,33}. \textbf{Data-driven methods} implicitly learn interactions among agents through neural networks. For instance, LSTM-based methods \cite{42,45} continuously refined social pooling patterns, while attention-based approaches \cite{10,31,37} dynamically adjusted agents' focal points in global interactions from a spatiotemporal integration perspective to optimize predicted trajectories. These methods primarily focused on straightforward interactions or direct responses between the target agent and its neighbors, often neglecting the higher-order, multidimensional influences exerted by surrounding agents on the target agent's intentions. Kim et al.~\cite{2} modeled agent intentions from 1st-order to N-order equally. However, when the number of agents was large, it may prioritized higher-order intentions, thereby diminishing the primary impact of lower-order intentions. Thus, the development of a multi-order fusion network that reinforces the dominant role of direct interactions while considering the secondary influence of higher-order transmissions on the target is essential.

\textbf{Associating future trajectory information.} Knowle- dge-driven methods face limitations due to the inherent flaws of unidirectional inference, which hinder effective association with future trajectory information. Data-driven methods exhibit two main shortcomings: first, agent movements are temporally dynamic, and relying solely on distance loss fails to capture these dynamic variations. Second, in generative model approaches \cite{14,32,38,39,40}, future trajectory information is implicitly involved in the training process without explicit guidance, resulting in poor interpretability. In safety-sensitive tasks such as autonomous driving, it is necessary to utilize meaningful and goal-oriented parameters and models.

\textbf{Efficiency.} Existing methods predominantly emphasize upstream tasks. For example, LSTM-based approaches \cite{12,23,35,36,41} modeled trajectory data in a serial temporal manner, while Transformer-based methods \cite{13,29,30,37} analyzed trajectory information through parallel processing. However, in downstream tasks, trajectories are often generated independently and sequentially. This method not only accumulates errors but also limits the efficiency of trajectory prediction. Therefore, establishing a trajectory parallel prediction model that accounts for global influences across future periods is a promising area for further exploration.

As illustrated in Fig.~\ref{fig1}, SocialMOIF is proposed to address the trajectory prediction task. A Multi-Order Intention Fusion model (MOIF) is introduced to guide the learning direction of the network. MOIF includes a first-order intention interaction layer and a higher-order intention interaction layer. The first layer focuses on direct interaction information between neighbors and the target agent, while the second layer addresses the indirect interaction information conveyed to the target agent after multiple interactions among neighbor targets. Inspiration is drawn from the squeeze theorem to design a trajectory distribution approximator to guide trajectories toward a distribution that closely aligns with ground truth and to enhance model interpretability. Simultaneously, RNN is incorporated to update the latent variables, achieving explicit supervision during training. The temporal dimension of decoded trajectories is aggregated to design a global trajectory optimizer. To the best of our knowledge, KANs \cite{5,6} are firstly introduced for the global optimization of trajectories, enabling more accurate and efficient parallel predictions. Furthermore, considering the travel directions of the agents, a distance-direction fused loss function is designed to comprehensively supervise the dynamic states of agents.

In summary, the main contributions of this paper are:
\begin{enumerate}
  \item[•] The proposal of SocialMOIF, which incorporates four novel designs:
  \begin{enumerate}
    \item[(1)] MOIF is introduced to comprehensively capture interaction information between the target agent and its neighbors;
    \item[(2)] A trajectory distribution approximator is developed to approximate and explicitly update the distribution of latent variables, guiding the distribution of trajectories;
    \item[(3)] A global trajectory optimizer is designed, and KANs are introduced for the first time to achieve parallel trajectory optimization across the global spatiotemporal domain;
    \item[(4)] A new direction-distance fused loss function is proposed for more effective supervision of agents' dynamic changes.
  \end{enumerate}
  \item[•] Comprehensive comparative experiments are conducted with existing high-quality models, demonstrating the effective and robust predictive capabilities of SocialMOIF.
\end{enumerate}
\section{Related works}
\textbf{Knowledge-driven methods.} Helbing et al.~\cite{56} first proposed the social force model, which describes pedestrian interactions through an energy potential field. Best et al.~\cite{47} introduced a Bayesian inference-based intention inference model, integrating observed movement trajectories with prior knowledge. EqMotion \cite{8} employed geometric analysis to convert human-human interaction trajectory analysis into an optimization problem by examining object geometry. Morris et al.~\cite{50} proposed an implicit Markov model for trajectory prediction, which is utilized for spatiotemporal probabilistic modeling of various types of pedestrian trajectories. A recent approach, socialcircle \cite{3}, incorporated angle-based light rule constraints into the trainable backbone to achieve interaction modeling of agent intentions.

Despite these advancements, these methods rely on knowledge from specific domains and therefore struggle to fully capture the complex agent intentions within social scenarios. Although attempts have been made to incorporate future trajectory information to overcome the limitations of unidirectional inference, the model's equations remain overly sensitive to common variations, leading to persistent challenges in trajectory prediction tasks \cite{19,20}.

\textbf{Data-driven methods.} LSTM-based methods model trajectory data temporally. Alahi et al.~\cite{45} employed a ``social pooling mechanism'' that considers only the intentions of agents within a grid while neglecting interactions with boundary agents. Gupta et al.~\cite{42} improved upon this by implementing a pooling mechanism that treats pedestrians in neighboring areas equally. Transformer-based methods enhanced temporal modeling approaches. Yu et al.~\cite{37} modeled and predicted trajectories using a spatio-temporal concatenation approach. Yin et al.~\cite{30} introduced optical flow to simulate agents' motion intentions. Graph neural networks were also utilized for trajectory prediction to model intention interactions among different nodes \cite{15,22,24,26,27,28}. Diffusion models \cite{9,17} were used to explore trajectory prediction based on randomness.

However, most of these methods primarily emphasize direct interactions between the target agent and its neighbors, often overlooking the higher-order influences that neighbors exert on the target agent. Furthermore, many of these approaches incorporate future trajectory information only at the network's output stage, supervising agents' trajectory changes solely through distance loss. Moreover, the limited interpretability of latent variable distributions in generative models can impede further development of these models.

In addition, the aforementioned methods mainly emphasize trajectory analysis and modeling of upstream tasks. For output in downstream tasks, results are generally produced independently in a sequential manner over time. This sequential processing introduces delays that may result in untimely responses, potentially leading to safety concerns. Furthermore, each future prediction is dependent on the state from the preceding time step, which may lead to accumulated errors, adversely affecting prediction accuracy.
\section{Approach}
\begin{figure*}[ht!]
	\centering 
	\includegraphics[width=\textwidth]{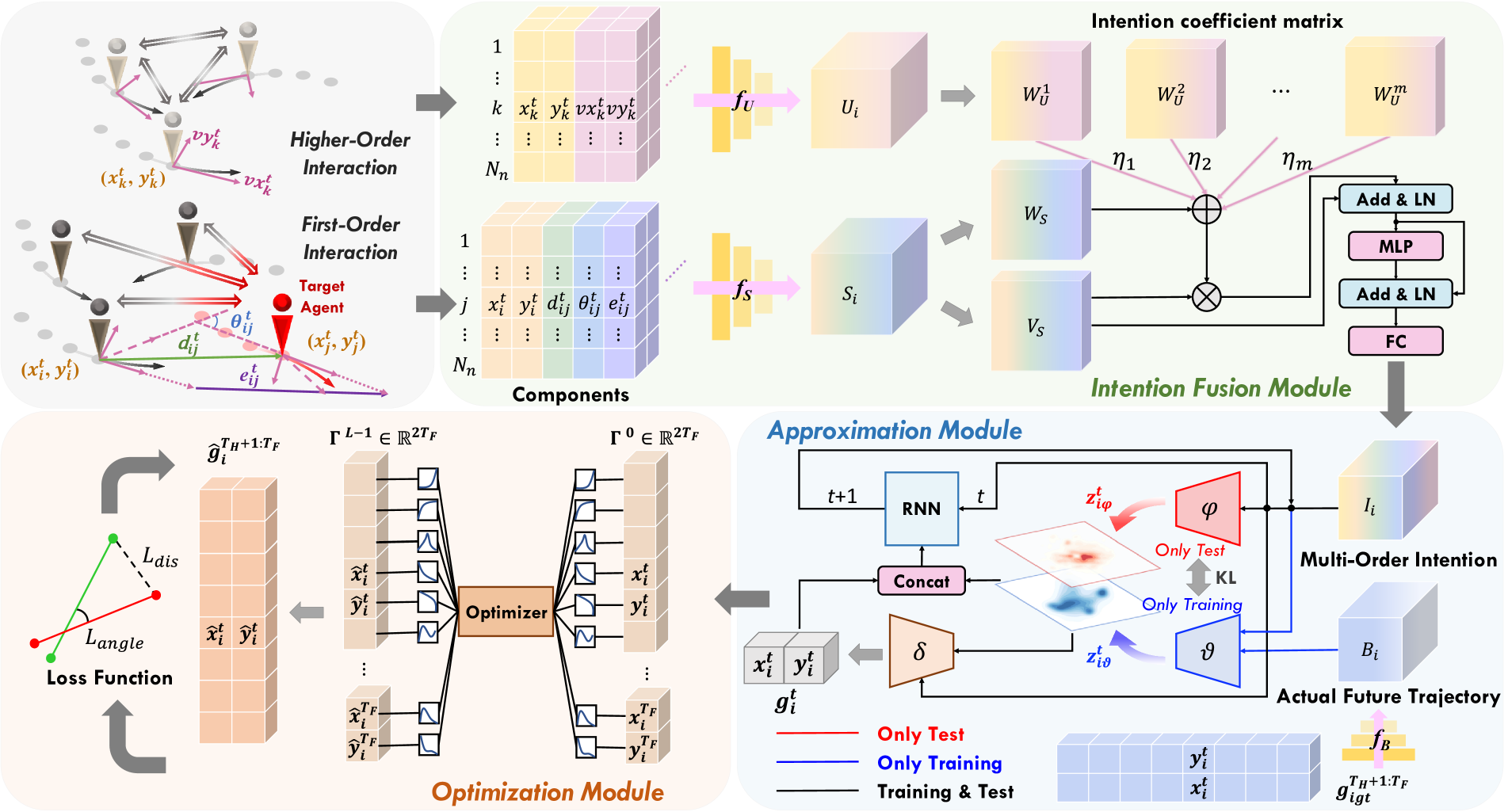}
	\caption{Computational pipeline of SocialMOIF. Firstly, the first-order interaction layer focuses on direct interactions between the target agent and its neighbors. The higher-order interaction layer addresses the interactions among neighbors. Multi-order intention is obtained by fusing the information from both layers. This multi-order intention is then combined with the actual future trajectory, with the distribution of the trajectory guided by an approximator. An optimizer is subsequently employed to enhance the decoded trajectory. Additionally, a fused distance-direction loss function supervises the training process. Finally, efficient parallel prediction is achieved.}
	\label{fig2}
\end{figure*}
The number of agents in the scenario is set to $\left(N_n+1\right)$. The trajectory of the target agent $i$ in history period $T_H$ is $g_i^{1: T_H}$, and $g_i^t$ is the two-dimensional spatial coordinate at moment $t$. The remaining agents are tentatively designated as neighbors, and the number is $N_n$. The trajectory of $i$ in the future period to be predicted is denoted as $\hat{g}_i^{T_H+1: T_F}$. We propose the multi-order intention fusion model in Section \ref{sec3_1}, the trajectory distribution approximator in Section \ref{sec3_2}, the global trajectory optimizer in Section \ref{sec3_3}, and the direction-distance fusion loss function in Section \ref{sec3_4}.

\subsection{Multi-order intention fusion model}
\label{sec3_1}
The model consists of two layers: a first-order intention interaction layer and a higher-order intention interaction layer. As shown in Fig.~\ref{fig2}, the first-order layer emphasizes the direct, dominant influence of neighbors on the target agent's intention. The higher-order layer accounts for the indirect and secondary influences on the target agent's intention that arise from interactions within the group of neighbors, thereby enabling a more comprehensive understanding of intentions. Based on the current and final history motion states of the agents, input components for the layers are first constructed, followed by the extraction and fusion of multi-order intention information.

\textbf{Higher-order intention interaction layer.} From a more fundamental perspective, the higher-order intention interactions within the group of neighbors can be decomposed into first-order intention interactions between individual neighbors. We define the intention influence of neighbor agent $a$ on neighbor agent $b$ as $w_{a \rightarrow b}$; for an agent itself, it is represented as $w_{a \rightarrow a}$. In a scenario with $N_n$ agents, the total number of such intentions is $\Omega$:
\begin{equation}\label{eq1}
  \Omega=\left(N_n-1\right) N_n+N_n=N_n^2
\end{equation}

To obtain the intentions, we extract the absolute positions and velocities of the $N_n$ neighbors corresponding to the target agent $i$. They are processed as a layer input component through the embedding layer $\boldsymbol{f}_S$ to obtain $U_i$.
\begin{equation}\label{eq2}
  U_i=\boldsymbol{f}_U\left(x_{1: N_n}^{1: T_H}, y_{1: N_n}^{1: T_H}, {v_x}_{1: N_n}^{1: T_H}, {v_y}_{1: N_n}^{1: T_H}\right)
\end{equation}

We capture the intention interactions of neighbors through multiple subspaces in parallel. $U_i$ is mapped into a learnable query vector $Q_U$, a key vector $K_U$, and a value vector $V_U$. The feature dimensions are flattened into M subspaces. For the m-th subspace, the overall intention coefficient matrix within neighbors can be expressed as:
\begin{equation}\label{eq3}
  W_U^m=\operatorname{SoftMax}\left(\frac{Q_U^m\left(K_U^m\right)^T}{d_M}\right)
\end{equation}

Thus, we obtain $w_{a \rightarrow b}=\left[W_U^m\right]_{a b}$. For this matrix, the number of elements is $\Omega$.

\textbf{The first-order intention interaction layer.} The layer input component of the target agent $i$ is processed through the embedding layer $\boldsymbol{f}_S$ to obtain $S_i$.
\begin{equation}\label{eq4}
  S_i=\boldsymbol{f}_S\left(x_i^{1: T_H}, y_i^{1: T_H}, d_{i \leftrightarrow 1: N_n}^{1: T_H}, \theta_{i \leftrightarrow 1: N_n}^{1: T_H}, e_{i \leftrightarrow 1: N_n}^{1: T_H}\right)
\end{equation}

We intensify the dominant role of direct interaction in components and focus on the design of interaction rules. For the present case, $d_{i \leftrightarrow j}^{1: T_H}$ is the distance between the target agent $i$ and neighbor $j$, and $\theta_{i \leftrightarrow j}^{1: T_H}$ is the velocity angle. For the final history, $e_{i \leftrightarrow j}^{1: T_H}$ is the final distance in the history period, assuming that after moment $t$, both the target agent $i$ and neighbor $j$ maintain the original velocity in the tangent direction $\overrightarrow{v x_i^t}, \overrightarrow{v x_j^t}$; then, $e_{i j}^t$ can be expressed as:
\begin{equation}\label{eq5}
  \lambda=\left|\left(\overrightarrow{d_{i j}^t} \cdot \overrightarrow{v_{i j}^t}\right) /\left\|\overrightarrow{v_{i j}^t}\right\|^2\right|
\end{equation}
\begin{equation}\label{eq6}
  e_{i j}^t=\left\|\overrightarrow{d_{i j}^t}+\min \left(\lambda, T_H-t\right) \cdot\left(\overrightarrow{v x_j^t}-\overrightarrow{v y_i^t}\right)\right\|
\end{equation}

Similarly, $S_i$ is mapped into learnable query vectors $Q_S$, key vectors $K_S$, and value vectors $V_S$. The matrix of the intention coefficients between the target agent $i$ and its neighbors is as follows:
\begin{equation}\label{eq7}
  W_S=\operatorname{SoftMax}\left(\frac{Q_S\left(K_S\right)^T}{d_S}\right)
\end{equation}

\textbf{Intention Fusion.} Higher-order interaction intentions are conducted to the target agent and combined with first-order interaction intentions to achieve multi-order intention fusion. During this fusion, the higher-order interaction intentions captured within each subspace exert varying effects when applied to the target agent. We set the corresponding trainable influence factor $\eta_m$ to denote this influence. The attention score $A_i$ is denoted as:
\begin{equation}\label{eq8}
  A_i=\left(\sum_{m=1}^M \eta_m W_U^m+W_S\right) V_S
\end{equation}
where $A_i$ is processed by two skip connections and a final fully connected layer to obtain the multi-order fused intention $I_i$.

\subsection{Trajectory distribution approximator}
\label{sec3_2}
We define the trajectory $g_{i g t}^{T_H+1: T_F}$ of the target agent $i$ in the future period. $B_i$ is obtained after feature enhancement through the embedding layer $\boldsymbol{f}_B$. By utilizing $I_i$ as the lower bound and $B_i$ as the upper bound, the latent variable distribution in the potential space is explicitly approximated. First, the probability distribution of the target agent $i$'s trajectory is initially modeled as:
\begin{equation}\label{eq9}
  \begin{aligned}
    &p\left(g_i^{T_H+1: T_F} \mid U_i, S_i\right)\\ 
    =&\prod_{t=T_H+1}^{T_F} p\left(g_i^{T_H+1+t} \mid g_i^{T_H+1: T_H+t-1}, U_i, S_i\right)
  \end{aligned}
\end{equation}

We sample the noise variables $\varepsilon \sim N(0,1)$ and reparameterize the latent variables that are approximated at $t$ \cite{21}:
\begin{equation}\label{eq10}
  \hspace{-6pt}z_{i \varphi}^t=u_{i \varphi}^t+\varepsilon \cdot \sigma_{i \varphi}^t \quad q_{\varphi}\left(\cdot \mid I_i^{t-1}, B_i^t\right) \sim N\left(u_{i \varphi}^t, \sigma_{i \varphi}^t\right)
\end{equation}
where $\varphi$ are parameters that need to be optimized during the training process. In the test mode, the latent variables $z_{i \vartheta}^t$ are only reasoned by $I_i$, and $\vartheta$ is also involved in the training process of the model.

We sample and decode the target agent's trajectories on the basis of latent variables and multi-order fused intention information.
\begin{equation}\label{eq11}
  g_i^t \sim \begin{cases}p_\delta\left(\cdot \mid z_{i \varphi}^t, I_i^{t-1}\right) & \text { if train } \\ p_\delta\left(\cdot \mid z_{i \vartheta}^t, I_i^{t-1}\right) & \text { if test }\end{cases}
\end{equation}
where $\delta$ are parameters that need to be optimized during the training process.

Concat the latent variables with the trajectories generated at the current time $t$, updating $I_i^{t-1}$ via $\boldsymbol{f}_{zg}$.
\begin{equation}\label{eq12}
  I_i^t= \begin{cases}\boldsymbol{f}_{z g}\left(\operatorname{concat}\left(z_{i \varphi}^t, g_i^t\right), I_i^{t-1}\right) & \text { if train } \\ \boldsymbol{f}_{z g}\left(\operatorname{concat}\left(z_{i \vartheta}^t, g_i^t\right), I_i^{t-1}\right) & \text { if test }\end{cases}
\end{equation}

In the next time step, the updated multi-order fused intention contains the information of the latent variables, which is then combined with $B_i^t$ to approximate the latent variables and update them.

By guiding the trajectory distribution through such a training process, the generated trajectories are explicitly represented by the latent variables as:
\begin{equation}\label{eq13}
  \begin{aligned}
    &p\left(g_i^{T_H+1: T_F} \mid U_i, S_i\right)\\ 
    =&\prod_{t=T_H+1}^{T_F} \int_{z_{i \varphi}^t} p\left(g_i^t \mid g_i^{T_H+1: t-1}, I_i^{t-1}, B_i^t, z_{i \varphi}^t\right)\cdot\\ 
    &p\left(z_{i \varphi}^t \mid g_i^{T_H+1: t-1}, I_i^{t-1}, B_i^t\right) d z_{i \varphi}^t
  \end{aligned}
\end{equation}

In test mode, $B_i^t$ is removed, and $\varphi$ is replaced with $\vartheta$.

\subsection{Global trajectory optimizer}
\label{sec3_3}
For efficient prediction, we introduce the KANs for the first time. We aggregate the $g_i^{T_H+1: T_F}$ time dimension to obtain $\Gamma^0 \in \mathbb{R}^{2 T_F}$ to facilitate global parallel optimization. The depth of the optimizer is $L$ layers, the number of nodes in the $\ell$-th layer is $m$, the number of nodes in the $\ell+1$-th layer is $n$, and the activation function matrix corresponding to the nodes in the $l$-th layer is $\Phi_\ell$. The optimization process for the $\ell \rightarrow \ell+1$ layer is as follows:
\begin{equation}\label{eq14}
  \Gamma^{\ell+1}=\Phi_\ell \Gamma^{\ell}=\left[\begin{array}{cccc}
  \varphi_{1,1}^{\ell} & \varphi_{1,2}^{\ell} & \cdots & \varphi_{1, m}^{\ell} \\
  \varphi_{2,1}^{\ell} & \varphi_{2,2}^{\ell} & \ldots & \varphi_{2, m}^{\ell} \\
  \vdots & \vdots & \ddots & \vdots \\
  \varphi_{n, 1}^{\ell} & \varphi_{n, 2}^{\ell} & \ldots & \varphi_{n, m}^{\ell}
  \end{array}\right] \Gamma^{\ell}
\end{equation}

Eventually, the predicted trajectory is reduced back to the original dimensional space to obtain the final trajectory over the entire prediction period at once.
\begin{equation}\label{eq15}
  \hat{g}_i^{T_H+1: T_F} \leftarrow \Gamma^L=\left(\Phi_L \circ \Phi_{L-1} \circ \cdots \circ \Phi_1 \circ \Phi_0\right) \Gamma^0
\end{equation}

Given the diversity of agent intentions, the process predicts multiple possible trajectories.

\subsection{Loss function}
\label{sec3_4}
Agents traveling in opposite or the same direction exert distinct effects on their respective movement states. Therefore, direction loss is incorporated to supervise the agent's dynamics during backpropagation.

For distance. The distance prediction error in future periods can be expressed as:
\begin{equation}\label{eq16}
  L_{dis}=\left\|\hat{g}_i^t-g_{i g t}^t\right\|
\end{equation}

For direction, the final moment of the historical period is introduced. Each predicted time step is related to the preceding time step to construct the direction vector, and subsequently, the angle between the predicted trajectory and the actual trajectory direction vector is analyzed.
\begin{equation}\label{eq17}
  L_{angle}=-\arccos \left(\frac{\left(\hat{g}_i^t-\hat{g}_i^{t+1}\right) \cdot\left(g_{igt}^t-g_{igt}^{t+1}\right)}{\left\|\hat{g}_i^t-\hat{g}_i^{t+1}\right\|\left\|g_{i g t}^t-g_{i g t}^{t+1}\right\|}\right)
\end{equation}

The objective of our model is to maximize the lower bound on the evidence for all time steps of the target distribution, joints (\ref{eq16}) and (\ref{eq17}), with a final loss function of:
\begin{equation}\label{eq18}
  \begin{array}{r}
    L=\sum\limits_{t=T_H+1}^{T_F}\bigg\{\mathbb{E}_{z_{i \varphi}^t \sim q_{\varphi}\left(\cdot \mid B_i^t, I_i^{t-1}\right)}\left[L_{dis}+L_{angle}\right]\\ 
    -KL\left[q_{\varphi}\left(z_{i \varphi}^t \mid B_i^t, I_i^{t-1}\right) \| q_{\vartheta}\left(z_{i \vartheta}^t \mid I_i^{t-1}\right)\right]\bigg\}
  \end{array}
\end{equation}

\section{Experiments}
\begin{table*}[ht!]
	\centering
	\caption{Comparison of Related Methods on ETH/UCY, NBA, SDD, and NuScenes (in meters). Every work was realized using best-of-20 predictions for the effect of ade/fde was selected.}
	\label{tab1}
	\tabcolsep 3pt
	\resizebox{\textwidth}{!}{
		\begin{tabular}{@{}c|cccccccccc|c@{}}
			\toprule
			\textbf{ETH/UCY}  & \textbf{FLEAM \cite{12}} & \textbf{ST-MR \cite{13}}  & \textbf{MTN \cite{30}}     & \textbf{MANTRA \cite{38}} & \textbf{Agent-former \cite{31}} & \textbf{BiTrap \cite{32}}       & \textbf{MemoNet \cite{11}} & \textbf{EqMotion \cite{8}}                                                  & \textbf{V2-Net-SC \cite{3,18}}              & \textbf{E-V2-Net-SC \cite{3,7}}             & \textbf{Ours}                             \\ \midrule
			Eth               & 0.66/1.21              & 0.53/0.69               & 0.58/0.86                & 0.56/0.97               & 0.58/0.86                     & 0.66/1.09                     & 0.51/0.69                & 0.36/0.54                                                                 & \textbf{0.23}/0.33 & \textbf{0.23/0.30} & 0.26/0.31                                 \\
			Hotel             & 0.37/0.64              & 0.20/0.31               & 0.19/0.57                & 0.17/0.32               & 0.15/0.24                     & 0.17/0.32                     & 0.14/0.20                & 0.12/0.16                                                                 & 0.11/0.16                                 & \textbf{0.10}/0.13 & \textbf{0.10/0.12} \\
			Univ              & 0.57/0.97              & 0.41/0.64               & 0.32/0.43                & 0.38/0.79               & 0.26/0.48                     & 0.26/0.46                     & 0.26/0.52                & 0.21/0.39                                                                 & 0.17/0.27                                 & 0.18/0.24                                 & \textbf{0.11/0.17} \\
			Zara01            & 0.46/0.74              & 0.26/0.38               & 0.24/0.37                & 0.24/0.41               & 0.19/0.32                     & 0.23/0.47                     & 0.21/0.36                & 0.16/0.27                                                                 & 0.13/0.18                                 & 0.13/\textbf{0.16}                                 & \textbf{0.10}/0.17 \\
			Zara02            & 0.37/0.69              & 0.18/0.24               & 0.15/0.21                & 0.18/0.34               & 0.16/0.25                     & 0.18/0.35                     & 0.15/0.30                & 0.11/0.19                                                                 & 0.12/0.21                                 & 0.11/0.16                                 & \textbf{0.09/0.14} \\
			Avg               & 0.48/0.85              & 0.32/0.45               & 0.30/0.49                & 0.31/0.57               & 0.27/0.43                     & 0.30/0.54                     & 0.25/0.41                & 0.19/0.31                                                                 & 0.15/0.23                                 & 0.15/0.20                                 & \textbf{0.13/0.18} \\ \midrule[0.75pt]
			\textbf{NBA}      & \textbf{FLEAM \cite{12}} & \textbf{ST-MR \cite{13}}  & \textbf{MTN \cite{30}}     & \textbf{SGNet  \cite{14}} & \textbf{EqMotion \cite{8}}      & \textbf{MemoNet \cite{11}}      & \textbf{BiTrap \cite{32}}  & \textbf{\begin{tabular}[c]{@{}c@{}}GroupNet\\ +CVAE \cite{15}\end{tabular}} & \textbf{V2-Net-SC \cite{3,18}}              & \textbf{E-V2-Net-SC \cite{3,7}}             & \textbf{Ours}                             \\ \midrule
			Rebound           & 1.92/2.21              & 0.94/1.73               & 0.84/0.97                & 0.80/1.78               & 0.86/1.21                     & 1.73/2.06                     & 0.86/1.88                & 0.63/0.84                                                                 & 0.54/0.79                        & 0.54/0.86                                 & \textbf{0.34/0.66} \\
			Scores            & 1.29/1.68              & 0.90/1.03               & 0.87/1.41                & 0.71/1.45               & 0.69/1.02                     & 1.35/1.87                     & 0.76/1.55                & 0.46/0.76                                                        & 0.50/0.84                                 & 0.47/0.78                                 & \textbf{0.30/0.56} \\ \midrule[0.75pt]
			\textbf{SDD}      & \textbf{FLEAM \cite{12}} & \textbf{SHENet \cite{16}} & \textbf{MemoNet \cite{11}} & \textbf{MANTRA \cite{38}} & \textbf{BiTrap \cite{32}}       & \textbf{MID \cite{17}}          & \textbf{Y-Net  \cite{33}}  & \textbf{NSP-SFM \cite{19}}                                                  & \textbf{V2-Net-SC \cite{3,18}}              & \textbf{E-V2-Net-SC \cite{3,7}}             & \textbf{Ours}                             \\ \midrule
			SDD               & 0.67/1.28              & 0.44/0.53               & 0.45/0.64                & 0.49/0.62               & 0.46/0.59                     & 0.38/0.73                     & 0.32/0.66                & 0.21/0.34                                                        & 0.26/0.36                                 & 0.23/0.35                                 & \textbf{0.17/0.24} \\ \midrule[0.75pt]
			\textbf{NuScenes} & \textbf{FLEAM \cite{12}} & \textbf{ST-MR \cite{13}}  & \textbf{MTN \cite{30}}     & \textbf{MANTRA \cite{38}} & \textbf{SGNet \cite{14}}        & \textbf{Agent-former \cite{31}} & \textbf{EqMotion \cite{8}} & \textbf{Y-Net \cite{33}}                                                    & \textbf{V2-Net-SC \cite{3,18}}              & \textbf{E-V2-Net-SC \cite{3,7}}             & \textbf{Ours}                             \\ \midrule
			NuScenes          & 2.02/3.14              & 1.84/2.46               & 1.64/2.12                & 1.47/2.01               & 1.28/2.09                     & 1.04/1.47                     & 1.24/1.98                & 1.12/1.71                                                                 & 1.06/1.59                                 & 1.04/1.64                        & \textbf{0.92/1.56} \\ \bottomrule
	\end{tabular}}
\end{table*}
\textbf{Datasets.} ETH \cite{53} and UCY \cite{54} datasets encompass over 1,600 pedestrian trajectories, with positions annotated at 0.4s intervals. SDD \cite{46} dataset is primarily utilized for pedestrian trajectory prediction and object tracking, containing about 19,000 pedestrian trajectories, with positions annotated every 0.4s. NBA dataset focuses on the 2015--2016 season \cite{44,48}. Two subsets, ``Rebounds'' and ``Scores,'' are extracted for benchmarking purposes. The movement intentions of the agents exhibit strong purposefulness and complex, hard-to-capture velocity variations. Consequently, the extracted trajectories encompass rich, highly nonlinear motion patterns. NuScenes \cite{34} is a dataset for autonomous driving, and the prediction challenge segmentation and configurations are adhered to.

\textbf{Metrics.} Best-of-20 predictions were used to compute Average Displacement Error (ADE) and Final Displacement Error (FDE) as primary metrics, with estimated Negative Log Likelihood (NLL) reported from the test set. See their detailed definitions in the Supplementary Material. 

\textbf{Implementation Details.} The model was trained on two NVIDIA RTX 4090 GPUs. Following the methodology in \cite{35,36}, a leave-one-out cross-validation approach was employed for dataset handling. The number of subspaces in the higher-order intention interaction layer was set to $M=6$, while the depth of the global trajectory optimizer was set to $L=3$. During inference, the best parameter model was loaded to predict the next 8 frames based on the previous 8 frames of history. For more implementation details of the training in the Supplementary Material.

\subsection{Quantitative analysis}
All SOTA performances across datasets were highlighted in \textbf{bold black} in Table~\ref{tab1}. In Eth scenario, crowds generally moved in parallel, resulting in weaker interactions among neighbors that conducted to target agent. E-V2-Net-SC \cite{3,7} achieved optimal performance in this scenario. However, the proposed method still achieved superior performance on ETH/UCY dataset, reducing ADE and FDE by 13.33\% and 10.00\% compared to E-V2-Net-SC, respectively. For NBA dataset, where interactions between agents were complex and frequent, the proposed method demonstrated exceptional performance. In Rebound sub-dataset, ADE and FDE decreased by 37.04\% and 16.46\%, respectively, compared to existing SOTA methods. In Scores sub-dataset, ADE and FDE were reduced by 34.78\% and 26.32\%, respectively, relative to the current SOTA methods. For SDD dataset, image segmentation techniques, as referenced in \cite{43}, were employed to capture pedestrian position information across eight scenario. Raw pixel coordinates were converted into spatial coordinates defined in meters for trajectory prediction tasks. NSP-SFM \cite{19} outperformed both V2-Net-SC and E-V2-Net-SC. Our method reduced ADE and FDE by 19.05\% and 29.41\%, respectively, compared to NSP-SFM. For NuScenes dataset, V2-Net-SC and E-V2-Net-SC showed similar performance levels. Our method reduced ADE and FDE by 11.54\% and 1.87\%, respectively, in comparison to their SOTA performances.

Detailed comparative experiments with NLL metric were carried out in the Supplementary Material. Furthermore, for wider validation and analysis, we applied directional component in the designed loss function to different models. See the Supplementary Material for details. 

\subsection{Qualitative analysis}
\begin{figure}[ht!]
  \centering 
  \includegraphics[width=\columnwidth]{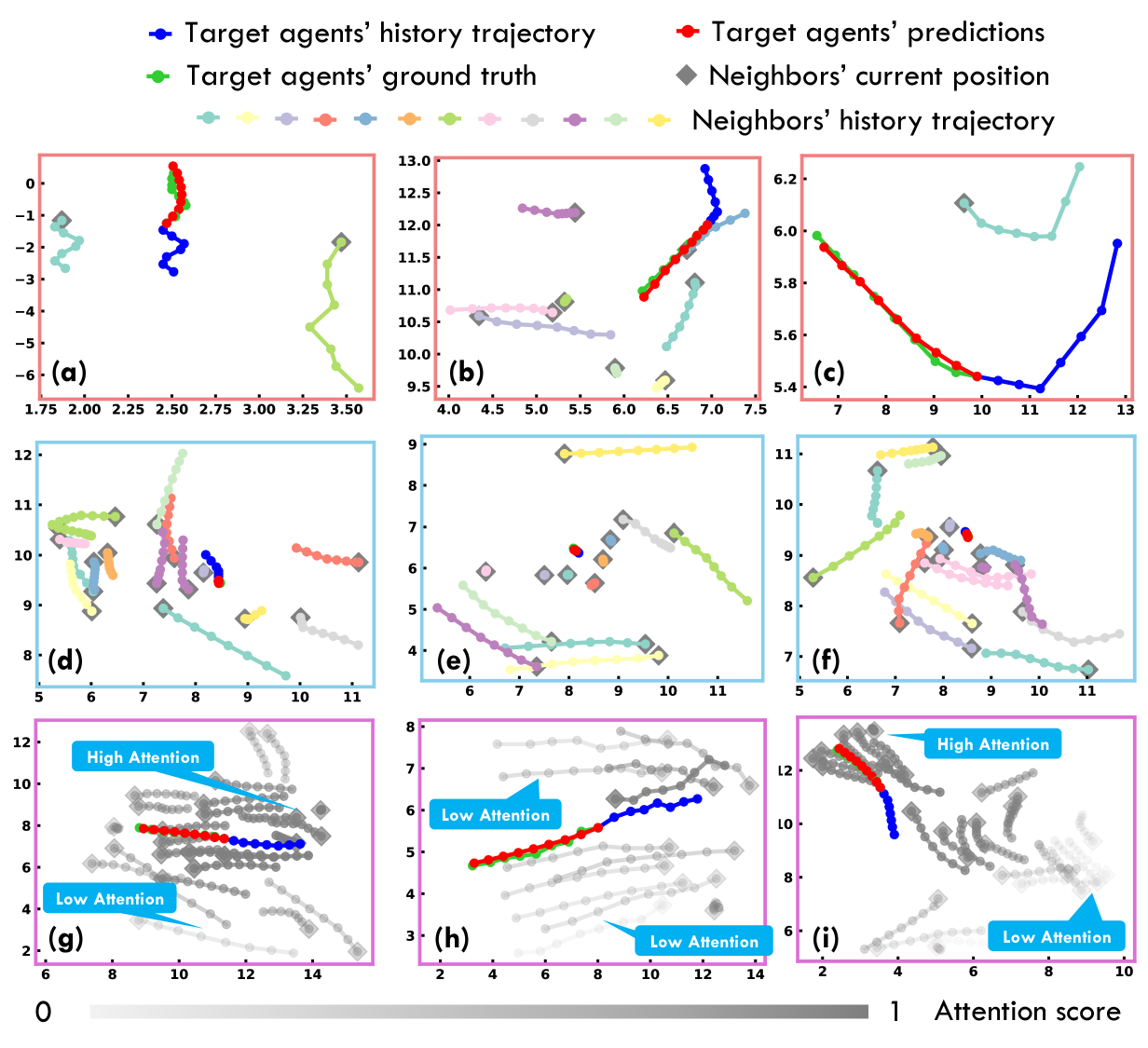}
  \caption{Representative cases. \textbf{\textcolor{lightcoral}{Row 1}} represented complex historical trajectory cases. \textbf{\textcolor{skyblue}{Row 2}} represented state maintenance cases. \textbf{\textcolor{orchid}{Row 3}} represented numerous neighbors cases.}
  \label{fig3}
\end{figure}

\begin{figure*}[ht!]
  \centering 
  \includegraphics[scale=0.52]{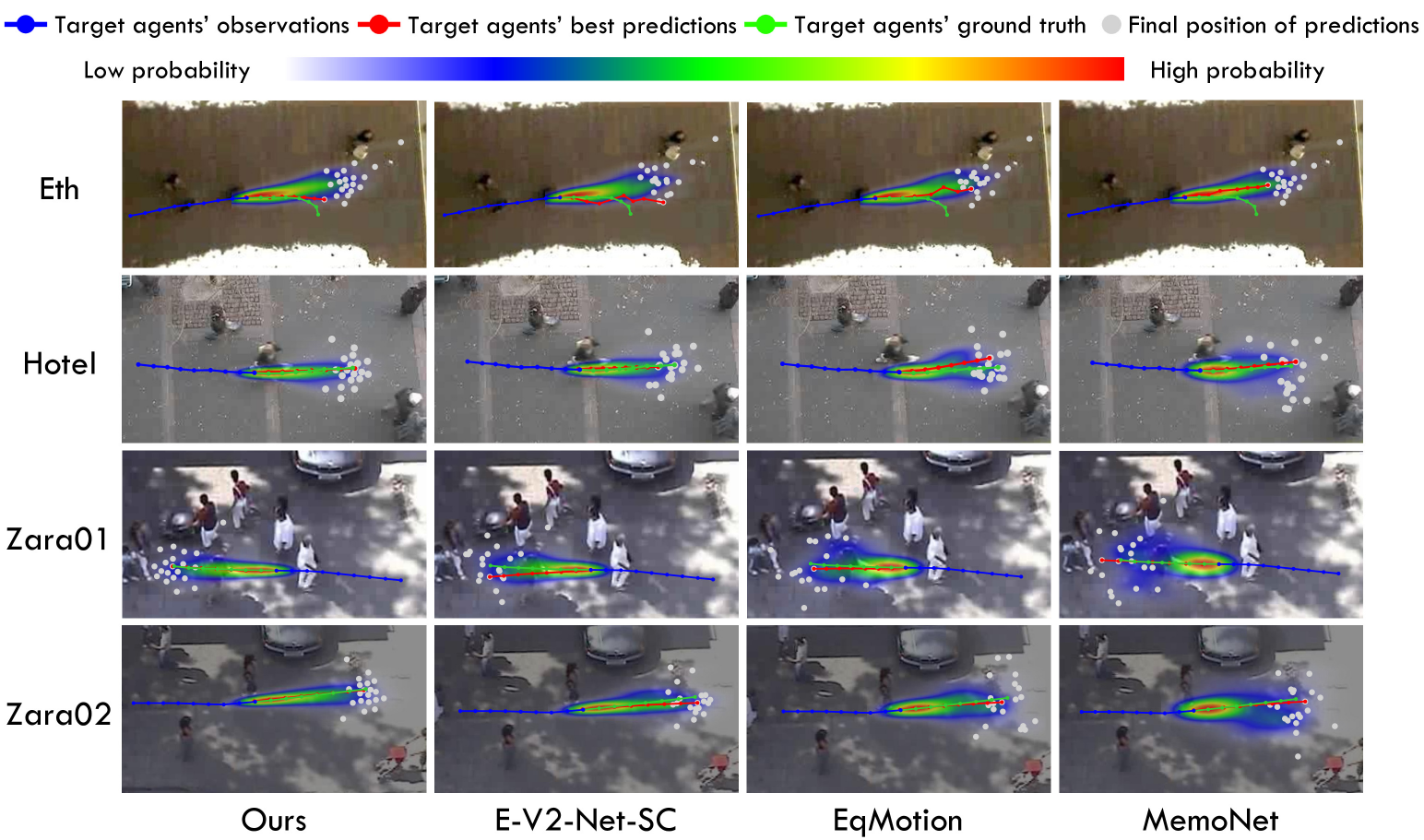}
  \caption{Predictions comparison of predictions in ETH/UCY. The comparison included heatmaps of the overall distribution of predictions and best-of-20 predictions of the related methods. }
  \label{fig6}
\end{figure*}
Unlike previous studies \cite{3,7,8,11,12,18,33}, the focus here was on conducting experiments involving historical trajectories in various social contexts. This enabled a more effective validation of the model's ability to extract and analyze multi-order fused intentions in Fig.~\ref{fig3}. In panel (a), the target agent continuously changed direction. Although the model displayed slight deviations at certain moments, these deviations remained within the acceptable range of directional changes. In panel (b) and (c), the target agent intended to turn, but the model successfully achieved accurate predictions. The target agent moved at a relatively steady speed in panel (d), adopting a conservative deceleration strategy in response to numerous neighbors who are stopping or remaining stationary. In panel (e) and (f), there were many stationary neighbors who may be gathered together in conversation. Consequently, the target agent maintained its original state. In panel (g) and (h), most neighbors moved in the same direction as the target agent, with closer neighbors receiving higher attention scores. This suggested that the target agent prioritized attention to closer neighbors, consistent with social norms. In panel (i), the directions of neighbors were complex, and their speeds varied. The target agent accounted for influences from neighbors in multiple directions while attempting to anticipate and avoid imminent collisions. Please see the qualitative analyses of NBA interactions in the Supplementary Material.

\begin{table*}[ht!]
  \centering
  \caption{Ablation studies of each component in our method on SDD and Nuscenes datasets. Notably, the positional information of the target agent was essential and was included in each set of ablation experiments. \textbf{P} represented the position of neighbors, \textbf{V} indicated their velocities, \textbf{D} denoted the distance between the target agent and its neighbors, and $\boldsymbol{\Theta}$ represented the angle between their velocities. \textbf{E} represented the final predicted distance between the target agent and its neighbors over historical periods, \textbf{I} indicated global historical social intent, \textbf{B} denoted future actual trajectories, \textbf{K} referred to the parallel prediction mode, and \textbf{A} signified directional information.}
  \label{tab4}
  \scriptsize
  \tabcolsep 12pt
  \begin{tabular}{c|ccccccccc|c|c}
  \toprule
  \textbf{Group} & \textbf{P} & \textbf{V} & \textbf{D} & $\boldsymbol{\Theta}$ & \textbf{E} & \textbf{I} & \textbf{B} & \textbf{K} & \textbf{A} & \textbf{SDD} & \textbf{Nuscenes} \\ \midrule
  \textbf{1}     & $\checkmark$ &   & $\checkmark$ &   &   & $\checkmark$ &   &   &   & 0.46/0.71    & 1.26/2.15         \\
  \textbf{2}     & $\checkmark$ & $\checkmark$ & $\checkmark$ &   &   & $\checkmark$ &   &   &   & 0.45/0.66    & 1.24/2.09         \\
  \textbf{3}     & $\checkmark$ & $\checkmark$ & $\checkmark$ & $\checkmark$ &   & $\checkmark$ &   &   &   & 0.41/0.56    & 1.18/1.97         \\
  \textbf{4}     & $\checkmark$ & $\checkmark$ & $\checkmark$ & $\checkmark$ & $\checkmark$ & $\checkmark$ &   &   &   & 0.35/0.43    & 1.12/1.84         \\
  \textbf{5}     & $\checkmark$ & $\checkmark$ & $\checkmark$ & $\checkmark$ & $\checkmark$ & $\checkmark$ & $\checkmark$ &   &   & 0.30/0.36    & 1.04/1.76         \\
  \textbf{6}     & $\checkmark$ & $\checkmark$ & $\checkmark$ & $\checkmark$ & $\checkmark$ & $\checkmark$ & $\checkmark$ & $\checkmark$ &   & 0.20/0.27    & 0.93/1.61         \\ \midrule
  \textbf{7}     & $\checkmark$ & $\checkmark$ & $\checkmark$ & $\checkmark$ & $\checkmark$ & $\checkmark$ & $\checkmark$ & $\checkmark$ & $\checkmark$ & \textbf{0.17/0.24}    & \textbf{0.92/1.56}      \\ \bottomrule
  \end{tabular}
\end{table*}

\begin{figure*}[ht!]
	\centering 
	\includegraphics[scale = 0.7]{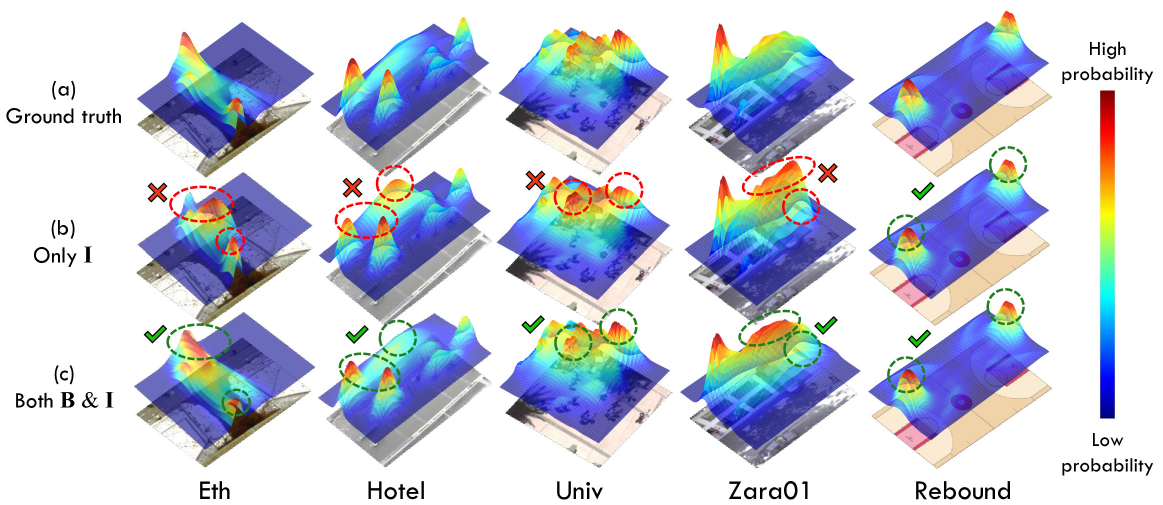}
	\caption{Distribution heatmaps. Every heatmap of future trajectory distribution was shown above the real scene image. (a) represented the actual distribution. (b) presented the experimental group containing only \textbf{I} component. (c) showed the group with both  \textbf{I} and \textbf{B} components.}
	\label{S_BI}
\end{figure*}
Comparisons were made between the proposed method and three previous methods in Fig.~\ref{fig6}. It was observed that our method produced more accurate predictions than the previous methods. In Eth, when facing neighbors to the left front, the model provided a prediction closest to the true trajectory. The predictions from E-V2-Net-SC and EqMotion exhibited slight oscillations, while MemoNet, by mimicking retrospective memory mechanisms, tends to maintain its original trajectory. By employing the squeeze theorem to approximate future true trajectories, the proposed model shows greater convergence in the overall distribution of predictions in all scenarios. In Hotel and Zara01, there were many pedestrians ahead of the target agent, and the diverging trajectory distributions from previous methods represent rather aggressive strategies, which conflicted with the safety requirements of autonomous driving.

\subsection{Ablation study}
In Table~\ref{tab4}, ablation experiments were conducted by sequentially adding components to the \textbf{PDI} baseline. Groups 1-4 focus on the multi-order intention fusion module. Group 1 includes only the neighbors' position and the distances between the target agent and the neighbors. As additional components, such as neighboring speed \textbf{V}, the angle between the target agent and the neighboring speed \textbf{$\boldsymbol{\Theta}$}, and the final predicted distance over the historical period \textbf{E}, were gradually incorporated, a substantial reduction in the final prediction error was observed. Specifically, ADE decreased by 23.91\% and FDE by 39.47\% on the SDD dataset, while ADE reduced by 11.11\% and FDE by 14.42\% on the Nuscenes dataset. Groups 4-5, 5-6, and 6-7 focus on the ablation of the approximator, the optimizer, and the angle loss term in the loss function, respectively. The results demonstrate that each component contributed positively.

\begin{figure}[ht!]
	\centering 
	\includegraphics[scale=0.23]{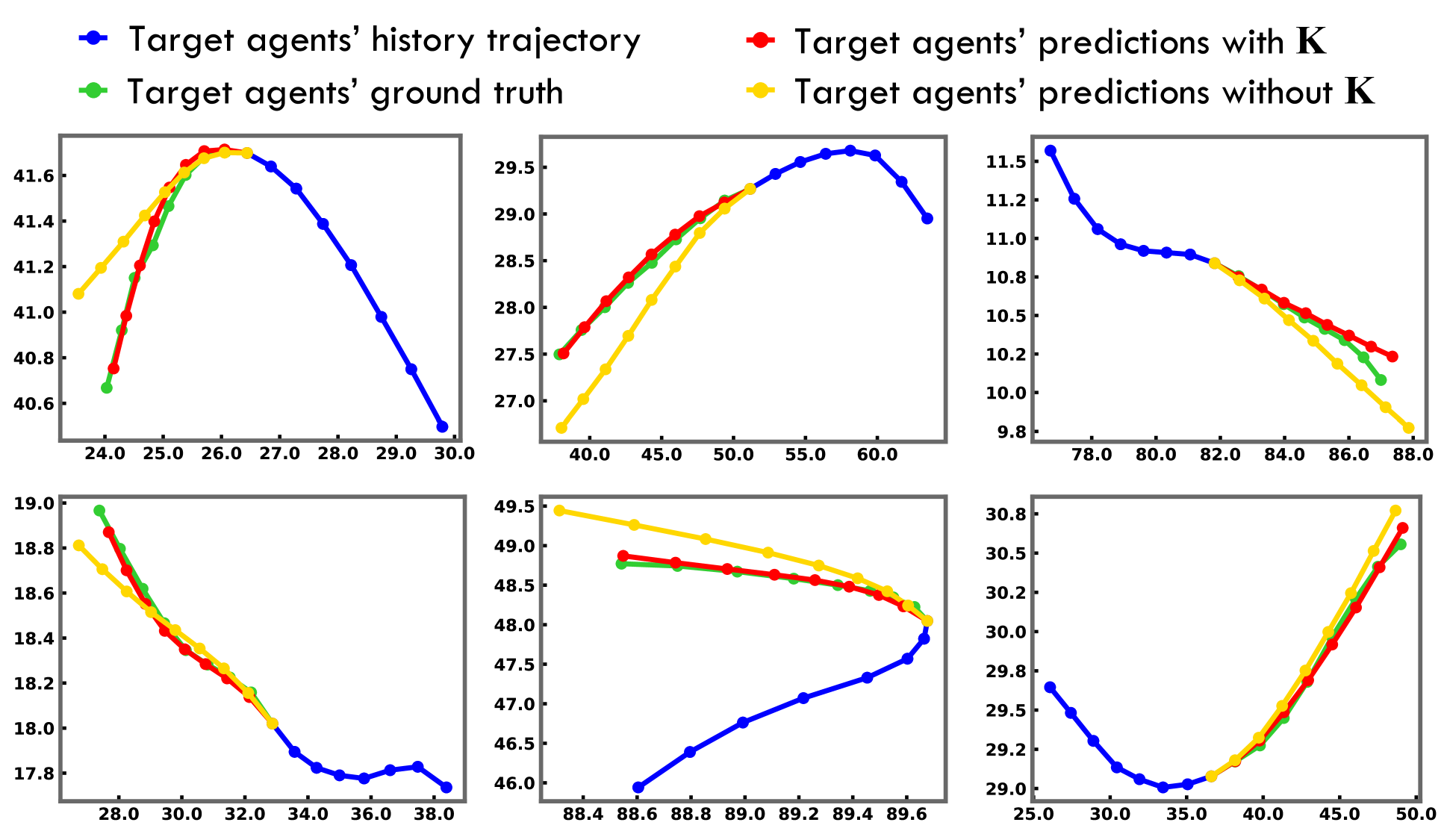}
	\caption{Representative cases about the optimization effects of component \textbf{K}. To ensure fairness, we selected best-of-20 predictions in groups 6 and 7 for comparison.}
	\label{fig9}
\end{figure}

To examine the influence of \textbf{B} and \textbf{I}, the trajectory distribution for several scenarios in the future periods was visualized in Fig.~\ref{S_BI}. It was evident that when both components \textbf{B} and \textbf{I} were included, the predicted trajectory distribution aligned more closely with the actual distribution. It can be found that in a strongly regular sports scenario such as Rebound, (b) and (c) achieved excellent predictions, while in open environments with complex social contexts, the effectiveness of the approximator was better demonstrated.

To validate the effectiveness of component \textbf{K} under global parallel optimization, six challenging cases were presented in Fig.~\Ref{fig9}. It was evident that predictions without component \textbf{K} exhibited increasing errors as the prediction time accumulated, whereas the parallel prediction mode demonstrated a closer result toground-truth.

\section{Conclusion}
Each human movement path is either directly or indirectly associated with neighbor agents. This study introduces SocialMOIF, which provides a more comprehensive representation of both direct and indirect intention interactions between the target agent and neighbor agents. SocialMOIF comprises four key designs: a multi-order intention fusion model, a trajectory distribution approximator, a global trajectory optimizer, and a novel loss function. Experimental results demonstrate that this approach performs competitively across various datasets and effectively approximates and explicitly updates trajectory distributions during training, thereby improving interpretability.

\section*{Acknowledgement}
This project is supported by National Natural Science Foundation of China (52202417), China Postdoctoral Science Foundation(2022TQ0155, 2022M721605), Open Project Program of State Key Laboratory of Virtual Reality Technology and Systems, Beihang University(VRLAB2023A02), Young Elite Scientists Sponsorship Program by CAST(2023QNRC001), Fundamental Research Funds for the Central Universities (NS2024030), Young Elite Scientists Sponsorship Program by JSTJ(JSTJ-2023-XH032).

{   
    \small
    \bibliographystyle{ieeenat_fullname}
    \bibliography{main}
}

\end{document}